\documentclass[conference]{IEEEtran}
\usepackage[T1]{fontenc}
\usepackage{times}
\usepackage[numbers,sort]{natbib}
\usepackage[bookmarks=true,hidelinks]{hyperref}
\usepackage{graphicx}
\usepackage{booktabs}
\usepackage{amsmath,amssymb}
\usepackage{array}
\usepackage{xcolor}
\let\labelindent\relax
\usepackage{enumitem}

\hypersetup{pdftitle={Where Does the Symmetry Break? Retrofitting a Sim-Trained RL Insertion Policy to Tilted Holes}, pdfauthor={Yi Huang, Xianming Liu, Zizhe Wang, Yao Zhu, Xiangyang Ji}}

\newcommand{\pih}{\pi_{\mathrm{h}}}

\begin{document}
\raggedbottom

\title{Where Does the Symmetry Break? Retrofitting a\\ Sim-Trained RL Insertion Policy to Tilted Holes}

\author{\IEEEauthorblockN{Yi Huang, Xianming Liu, Zizhe Wang, Yao Zhu, Xiangyang Ji}}

\maketitle

\begin{abstract}
\boldmath
Reinforcement learning policies for sub-millimeter insertion are costly to retrain for every hole tilt.
A cheap \emph{retrofit} expresses a policy's pose observations and actions in a frame attached to the hole, where a tilted hole looks vertical.
This rotates the task but not the robot: the base stays on the table and gravity still points down.
We ask where this symmetry breaks for a grasp--carry--insert policy trained in simulation for a real cell, at a vertical hole and with a load on its hand.
In simulation, started above the hole, the retrofitted policy keeps nearly all its success at a hole tilted $15^\circ$ away from the robot; its success falls from 91\% at the vertical hole to 45\% at $45^\circ$.
Correcting what the frame leaves unrotated---the joint angles the policy reads, the arm's stiffness and damping, its joint limits, and the direction of the loads on the hand---raises success at $45^\circ$ to 87\%.
Split by Shapley values, the arm accounts for over half of the drop in log-odds at $30^\circ$ and $45^\circ$ away from the robot and the loads for a quarter to a third; sideways, the symmetry also breaks after the part is released.
Fine-tuning in the hole frame for a fifth of the original training lifts full-task success at three tilts away from the robot from 12--44 to 73--76\%.
Planned experiments will use the corrections to predict, before retraining, what redesigned policies lose at a tilt, and test the drops and the fine-tuning gain on a real arm.
\end{abstract}

\IEEEpeerreviewmaketitle

\section{Introduction}

Reinforcement learning (RL) in simulation yields insertion policies for sub-millimeter clearance~\cite{narang2022factory} that transfer to real robots~\cite{tang2023industreal,tang2024automate}.
Such policies are costly to train and tied to the hole tilt seen in training. Ours, the \emph{source policy}, took about three days of simulation; trained at a vertical hole with the tool pointing down, it cannot insert at holes tilted $15^\circ$ to $45^\circ$.
A cheap fix treats rotating the hole as a symmetry of the task: it expresses the policy's pose observations in a frame attached to the hole (the \emph{hole frame}), where a tilted hole looks vertical, and rotates its actions back, keeping its weights.
We ask where this symmetry breaks for a policy not designed for it, and what each break costs.

\begin{figure}[t]
\centering
\includegraphics[width=\columnwidth]{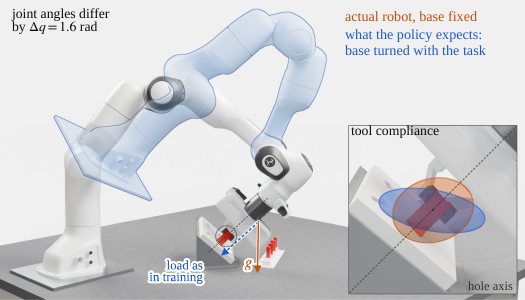}
\caption{\textbf{The hole frame rotates the task, not the robot.} A simulated episode at $-45^\circ x$, rendered with the CAD of the real cell. The arm (solid) reaches the tool pose, relative to the hole, that the policy learned at the vertical hole, but with the base fixed. The policy expects that configuration turned with the task (translucent blue): its joint angles differ by 1.6\,rad, gravity on the hand (orange) acts $45^\circ$ off the hole axis instead of along it (blue), and the tool's compliance (inset: ellipsoids $x^\top C_p^{-1}x=\mathrm{const}$, actual orange, expected blue) differs in shape and is 1.6 times as large along the hole axis. The arm stays clear of its joint limits here.}
\label{fig:overview}
\end{figure}

Policies with observations, actions and controller in the tool frame, relative to the goal, insert into tilted holes on real arms, provided the robot can execute the rotated task~\cite{seo2024gic,seo2025equicontact}.
IndustReal and AutoMate policies command isotropic task-space impedance but, like ours, act in the base frame and read the joint angles~\cite{tang2023industreal,tang2024automate}.
Wrapped in the hole frame, such a policy sees its pose inputs and actions turn with the hole, but not the robot or gravity (Fig.~\ref{fig:overview}).
With the base fixed, the arm reaches the tool pose the policy expects in another configuration, so four factors differ from training: the \emph{joint angles} the policy reads, the tool's \emph{stiffness and damping} under joint impedance, the margin to \emph{joint limits} and singularities, and the direction of the \emph{loads} on the hand, mainly a payload, relative to the hole.
Isotropic task-space impedance removes only the second; tool-frame policies also remove the first. How much each factor costs a retrofitted policy has not been measured.

We measure it in simulation, where the factors can be separated and the base can even be turned with the task.
We correct each factor at the tilted hole to see how much of the drop in success it recovers, and inject it at the vertical hole to see how much of the drop it reproduces.
Over 17 tilts, the retrofitted policy loses more the further the arm's joint angles move from those in training. Preliminary results (Sec.~\ref{sec:results}) support H1 and H2 below; Table~\ref{tab:crit} states the criteria, and planned experiments test all three hypotheses (Sec.~\ref{sec:plan}).

\textbf{H1 (where the symmetry breaks).} From above the hole, the retrofitted policy's success falls as the arm's joint angles move away from those in training, across tilts and across board placements at a fixed tilt; correcting the four factors recovers most of the drop at tilts away from the base.

\textbf{H2 (fine-tuning in the hole frame).} On the full task, fine-tuning in the frame for a fifth of the original training brings tilts away from the base to similar insertion rates, at least as high as that of the source fine-tuned the same way at the vertical hole and above that of residual RL.

\textbf{H3 (corrections predict designs).} A policy retrained without joint-angle inputs, with isotropic task-space impedance or without the payload avoids the share of the drop that the matching correction recovers.

Our contributions are:
\begin{itemize}[leftmargin=*,nosep]
\item a correct-and-inject protocol that measures the cost of each factor a retrofit leaves unrotated;
\item the finding that the symmetry breaks at the arm and in the loads: with both corrected, the retrofitted policy comes within 4 points of the vertical hole at $30^\circ$ and $45^\circ$ away from the base;
\item the result that fine-tuning in the hole frame for a fifth of the original training brings the insertion rate at tilts away from the base to at least that of the source fine-tuned the same way at the vertical hole, and about 19 points above residual RL at $-30^\circ x$.
\end{itemize}

\section{Related Work}

\textbf{Goal-frame and equivariant policies.}
Policies generalize over object poses through equivariant networks~\cite{wang2022so2,wang2024equidiff,yang2024equibot} or by acting in a task frame~\cite{mason1981compliance,calinon2016tutorial,rana2025affordance,zhao2025hep}. Seo~et~al.\ learn the gains of a tool-frame impedance controller, so the closed loop transforms with the goal~\cite{seo2024gic}; with ground-truth frames, EquiContact inserts on a real arm into holes rotated up to $135^\circ$~\cite{seo2025equicontact}. Canonicalization and equi-tuning make pretrained networks equivariant by transforming inputs and outputs~\cite{kaba2023canonicalization,mondal2023equivariant,basu2023equituning}; reuse through such a map~\cite{taylor2007intertask} needs dynamics sharing the symmetry, as in MDP homomorphisms~\cite{ravindran2003relativized,soni2006homomorphisms,vanderpol2020mdp}.

\textbf{When the robot breaks the symmetry.}
Joint limits, singularities~\cite{park2025steering,seo2025equicontact} and gravity break task symmetries, which caps an equivariant policy's accuracy~\cite{wang2023extrinsic}; partially~\cite{chang2026partial} and approximately~\cite{park2025steering} equivariant RL relax them. Proprioceptive inputs such as joint angles limit imitation policies' spatial generalization~\cite{zhao2025statefree}; joint impedance makes tool stiffness and damping configuration-dependent~\cite{salisbury1980active,ott2008cartesian}. Unlike studies of scene factors~\cite{xie2024decomposing}, we correct and inject arm and load factors. Closest to ours, Seo et~al.\ and EquiContact build the symmetry into the policy and its controller~\cite{seo2024gic,seo2025equicontact}, and IndustReal and AutoMate remove the configuration dependence of the tool's stiffness through task-space impedance~\cite{tang2023industreal,tang2024automate}. We instead take a policy trained without the symmetry in mind and measure what each robot-side factor costs it at a tilt, which predicts which of these design choices matter, as our planned experiments test (H3).

\textbf{Adapting assembly policies.}
A new hole pose can be covered by randomizing it in training ($\pm25^\circ$ in~\cite{liu2025genpih}), by moving demonstration segments to it~\cite{mandlekar2023mimicgen,garrett2024skillmimicgen} and refining behavior-cloned policies with RL~\cite{zhou2025reinforcegen}, or by a residual policy~\cite{johannink2019residual} on a frozen one~\cite{ankile2025resip}. SRSA predicts which library skill transfers best to a new part~\cite{guo2025srsa}.

\section{Setting}
\label{sec:setting}

\begin{figure*}[t]
\centering
\includegraphics[width=\textwidth]{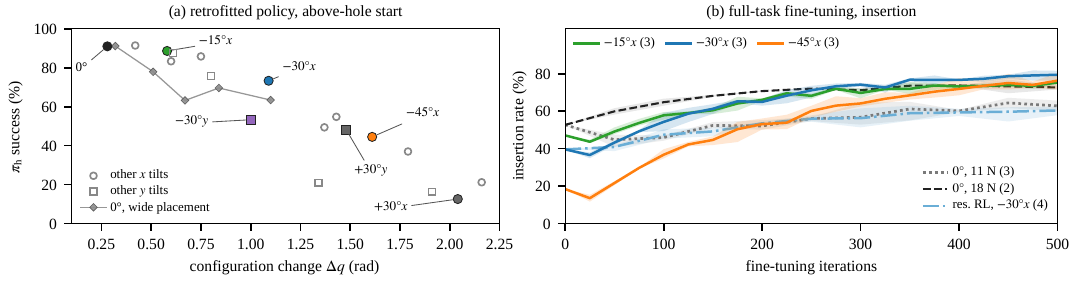}
\caption{\textbf{Success falls with the configuration change; fine-tuned in the frame, the $-x$ tilts at least match the vertical hole.} (a)~Success of $\pih$ above the hole; diamonds: $0^\circ$ with the board placed widely, binned by each episode's $\Delta q$. (b)~Insertion rate during full-task fine-tuning, mean and range over runs: the $-x$ tilts in the frame, residual RL at $-30^\circ x$, and the source at $0^\circ$ with the pressing-force threshold at 11 or 18\,N; iteration 0 is on the fixed board.}
\label{fig:tilt}
\end{figure*}

\textbf{Task.}
All experiments here run in Isaac Lab~\cite{mittal2023orbit,mittal2025isaaclab}, with a 7-DoF Franka Panda arm, gravity-compensated and driven by a joint impedance controller. From the arm's rest pose, the policy grasps a 10\,g L-shaped part from a rack, carries it, inserts it into a matching blind hole (20\,mm deep, 0.22\,mm clearance) in a board, and releases it, ending its control. An episode succeeds if the part then settles at least 12\,mm deep along the hole axis within 1\,s; it fails at 30\,s or early, e.g., if the wrist force exceeds 30\,N.

\textbf{Source policy.}
The source policy, an LSTM trained with PPO~\cite{schulman2017ppo}, observes the tool's pose and velocity relative to a hole estimate, the part's estimated pose, a wrist-force estimate, gripper state, previous action and joint angles, and outputs position and yaw steps of the tool, which stays pointing down, and a gripper command. A latched \emph{insertion mode}, flagged in the observation once the part is near the hole, replaces free-space moves by small insertion steps. The source was trained for our real cell, where the board is not clamped and the part can stick partway into the hole; we retrofit it as is. Four of its training randomizations matter below: a downward \emph{payload} of 0.5--4\,N on the hand, standing in for the real arm's uncompensated load and zeroed out of the force reading at the start; a vertical \emph{sticking resistance} that, in some episodes, stops the held part partway down until it is released; \emph{bumps} to the free board in some full-task episodes; and errors in the periodically refreshed hole estimate. Training penalizes the world-vertical wrist force above 11\,N.

\textbf{Tilted holes.}
A wedge under the board tilts the hole axis about the board's $x$ axis ($-60^\circ$ to $+30^\circ$) or $y$ axis ($\pm15^\circ$ to $\pm45^\circ$) and raises the mouth. Our main cases are $-15^\circ x$, $-30^\circ x$ and $-45^\circ x$ (the \emph{$-x$ tilts}, mouth away from the base) and $-30^\circ y$ (sideways). A free wedge would be pushed away, so fine-tuning and evaluations fix the board, which was free in the source's training; the $0^\circ$ reference runs on the same fixed board and, for H2, is also fine-tuned on it. It is simulated in two setups; each condition is compared with the $0^\circ$ run of its own setup (Table~\ref{tab:decomp}).

\textbf{Retrofit.}
The \emph{retrofitted policy} $\pih$ runs the source's network unchanged on a virtual vertical task. Let $R$ be the tilt, of known angle, about the board's estimated $x$ (or $y$) axis, and $o$ the mouth estimate. Positions $p$ are mapped to $R^\top(p-o)+o$, and orientations and velocities are rotated by $R^\top$. The target that the policy's steps produce, including the pointing-down orientation, is rotated back by $R$. The wrist force (tool frame), joint angles and previous action pass unchanged. The frame switches on during the carry near the mouth and stays on while the part is held. As a contrast in design, we also retrofit a Factory peg-insertion policy~\cite{narang2022factory} (0.11\,mm clearance) with isotropic task-space impedance, no joint-angle inputs and no gravity or payload on arm or peg; of the four factors, only the joint limits apply to it.

\textbf{Start conditions and evaluation.}
\emph{Above-hole starts} hold the part just above the mouth, near its axis, with the frame and insertion mode on and no bumps; they isolate the insertion (H1). \emph{Full-task starts} begin from the rest pose (H2); we report full-task success and the \emph{insertion rate}, the success rate of episodes that enter the insertion mode. Each evaluation runs 512 episodes with the mean action and random board and rack poses as in training; Tables~\ref{tab:zs} and~\ref{tab:decomp} pool four evaluation seeds (2048 episodes) unless noted. A fine-tuning iteration is one PPO update on 2048 environments of 128 steps; the source trained for about 2560.

\section{What the Frame Cannot Rotate}
\label{sec:loop}

This section states when the retrofit is exact, which defines the four factors. Inverse kinematics turns the policy's target tool pose into joint targets $q_d$, and the controller applies
\begin{equation}
\tau=K_q(q_d-q)-D_q\dot q,
\label{eq:ctrl}
\end{equation}
$q$ being the joint angles, with diagonal $K_q$, randomized per episode, and $D_q$. Near the target the tool behaves approximately as a spring of compliance $C$ and a damper $B$,
\begin{equation}
C(q)=JK_q^{-1}J^\top,\qquad B(q)=(JD_q^{-1}J^\top)^{-1},
\label{eq:cb}
\end{equation}
$J(q)$ being the tool Jacobian in the base frame. Rotated by $R$ about the mouth, a source trajectory with joint angles $q(t)$ is reached at the tilted hole with joint angles $q'(t)$. From above-hole starts, the retrofitted loop is the source's loop rotated by $R$, and succeeds as often, if along every such trajectory
\begin{itemize}[leftmargin=*,nosep]
\item \emph{joint angles:} the joint-angle input is $q$, not $q'$;
\item \emph{stiffness and damping:} the impedance turns with the task, approximately $C(q')=AC(q)A^\top$ and $B(q')=AB(q)A^\top$, $A=\mathrm{diag}(R,R)$;
\item \emph{joint limits:} the arm stays clear of joint limits and singularities;
\item \emph{loads:} the loads on the hand are $Rw$, $w$ being the payload plus the held part's weight at the vertical hole;
\end{itemize}
and nothing else differs. Other differences, such as the arm's inertia and friction, gravity on the released part, the sticking resistance and hole-estimate errors, stay uncorrected.

We describe each start by three quantities: the distance $\Delta q=\|q'-\bar q_0\|$ of its joint angles $q'$ from the mean $0^\circ$ start $\bar q_0$; the compliance $c_\parallel=n^\top C_p n$ along the hole axis $n$, $C_p$ being the translational part of $C$ (both in Table~\ref{tab:zs}); and the payload's sideways deflection, the part of $C_p$ times the payload across $n$, minus its $0^\circ$ mean.

\section{Methodology}
\label{sec:method}

Table~\ref{tab:crit} gives each hypothesis's criterion. Shares and gains are in log-odds (logits), $\ell=\ln\frac{s}{1-s}$ for a success rate $s$; differences of rates are in points. The \emph{drop} at a tilt is $\ell_0-\ell_R$, the logit at $0^\circ$ minus the tilt's (above-hole starts). Intervals $[a,b]$ are 95\% bootstrap intervals over episodes, paired where conditions share start draws; for Table~\ref{tab:decomp}'s cells, we resample evaluation seeds, then episodes. Logistic regressions have a payload covariate and an intercept per evaluation seed.

\textbf{H1, configuration.}
We evaluate $\pih$ from above the hole at the 17 tilts (Fig.~\ref{fig:tilt}a) and report the Spearman correlation $\rho$ of success with the median $\Delta q$. Across tilts, the configuration and the load direction change together. Moving the board at a fixed tilt changes the configuration but not the loads in the hole frame, so at $0^\circ$, $-15^\circ x$, $-30^\circ y$ and $-45^\circ x$ we also place it within an 18\,cm disk and $\pm60^\circ$ in yaw and regress success on each episode's $\Delta q$. Unreachable starts count as failures in $\rho$ and are excluded from the regressions. To see which part of the configuration change matters, we also add $c_\parallel$ and the payload's sideways deflection as covariates.

\textbf{H1, corrections.}
At $-30^\circ x$, $-45^\circ x$ and $-30^\circ y$ we correct the factors one at a time and together:
\begin{itemize}[leftmargin=*,nosep]
\item \emph{joint angles:} the policy reads the joint angles $q_v$ of a virtual robot whose base is turned with the task, at the same tool pose;
\item \emph{stiffness and damping:} a torque added to the control law \eqref{eq:ctrl},
\begin{equation}
\tau_c=\lambda J^\top\big(A\,C(q_v)^{-1}A^\top-C(q')^{-1}\big)J\,(q_d-q'),
\label{eq:tauc}
\end{equation}
with $J=J(q')$, moves the tool's stiffness halfway ($\lambda=0.5$) toward that of $q_v$; with \emph{matched damping}, it also moves the damping $B$ (both stable at all three tilts);
\item \emph{joint limits:} relaxed by 0.3\,rad;
\item \emph{loads:} the loads on the hand are turned with the task, the part's weight only while it is held.
\end{itemize}
A correction with logit $\ell_{\mathrm{cor}}$ recovers the share $(\ell_{\mathrm{cor}}-\ell_R)/(\ell_0-\ell_R)$ of the drop. In reverse, we inject at $0^\circ$ a tilt's joint angles, its stiffness and damping (halfway, as corrected) or its loads, also with the released part's weight; an injection with logit $\ell_{\mathrm{inj}}$ reproduces the share $(\ell_0-\ell_{\mathrm{inj}})/(\ell_0-\ell_R)$ of the tilt's drop. Correcting and injecting agree when a factor costs the same at the tilt as it does alone at the vertical hole.

Two references, possible only in simulation, turn the base with the task instead. \emph{Base turned} runs at $0^\circ$ with the tilt's loads, including the released part's weight, and part of its hole-estimate errors; for a gravity-compensated arm, this removes all other differences, including those of the joint angles, impedance and limits. \emph{All rotated} also sets the loads as at $0^\circ$, leaving only those errors. As controls, we turn the payload alone and the loads the other way, move stiffness and damping a quarter of the way the other way ($\lambda=-0.25$), and rerun the main conditions at $-30^\circ x$ and $-45^\circ x$ without the payload. Shapley values~\cite{shapley1953value} split the joint share of the joint-angle, stiffness-and-damping and load corrections over their eight combinations (relaxing the limits changes no episode). The Factory policy is evaluated at tilts matched by direction to the robot, also with relaxed limits.

\textbf{H2.}
We fine-tune $\pih$ with PPO for 500 iterations at the three $-x$ tilts, three runs each. As the reference, we fine-tune the source at $0^\circ$ the same way (three runs). All runs use the source's rewards; 15\% of training episodes start at the rest pose, the others later in the task. The rewards' pressing-force penalty acts on the world-vertical force, which is axial at $0^\circ$; at $-30^\circ x$, it is about as large as the axial force in contact. A second $0^\circ$ reference (two runs) raises the penalty's threshold to 18\,N, by the ratio of $c_\parallel$ at $-30^\circ x$ to $0^\circ$, and keeps the force limit. The baseline at $-30^\circ x$ is residual RL~\cite{johannink2019residual,ankile2025resip}, which trains, in the frame, a correction added to the frozen $\pih$'s actions. It runs in two settings, two runs each: the correction bounded by twice its exploration noise, with a lower critic learning rate than fine-tuning; and, retuned, bounded by three times the noise, with fine-tuning's critic rate.

\begin{figure}[t]
\centering
\includegraphics[width=\columnwidth]{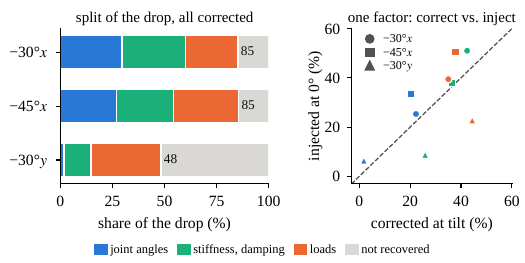}
\caption{\textbf{At $-30^\circ x$ and $-45^\circ x$, the arm takes 54--60\% of the drop and the loads 25--31\% (H1).} Loads: all loads while held. Left: share recovered by correcting the joint angles, stiffness and damping, and loads together, split by Shapley values; numbers: total. Right: one factor's share, corrected at the tilt against injected at $0^\circ$ (relative to the 89.7\% run; injected stiffness and damping: 1024 episodes); dashed: equal.}
\label{fig:decomp}
\end{figure}

\textbf{H3.}
We will retrain the source in four variants and decompose each as in Table~\ref{tab:decomp} and Fig.~\ref{fig:decomp}. Above the hole at $-30^\circ x$, $-45^\circ x$ and $-30^\circ y$, H3 predicts a variant's logit at the tilt as $\tilde\ell_0-(1-\phi)(\ell_0-\ell_R)$, $\tilde\ell_0$ being its logit at $0^\circ$ and $\phi$ the share of the source's drop that the matching correction recovers (Table~\ref{tab:decomp}). The matching correction is the joint-angle one for the variant without joint-angle inputs; stiffness with matched damping, as in Table~\ref{tab:decomp}, for the variant with isotropic task-space impedance; the payload turned alone for the variant without the payload; and these corrections together for the variant with all three changes.

\begin{table}[t]
\centering
\caption{Hypotheses, criteria and preliminary outcome}
\label{tab:crit}
\footnotesize
\setlength{\tabcolsep}{2.5pt}
\begin{tabular}{@{}>{\raggedright\arraybackslash}p{0.13\columnwidth}>{\raggedright\arraybackslash}p{0.57\columnwidth}>{\raggedright\arraybackslash}p{0.26\columnwidth}@{}}
\toprule
 & Criterion & Outcome \\
\midrule
H1, config. & With the board placed widely at $0^\circ$, $-15^\circ x$, $-30^\circ y$ and $-45^\circ x$, the slope of success on each episode's $\Delta q$ and its interval lie below 0 at each tilt. & Holds: slopes $-1.5$ to $-0.9$ logits per rad. \\
\addlinespace
H1, corr. & All four corrections recover more than half of the drop at $-30^\circ x$ and $-45^\circ x$ (point estimates); the Factory policy loses only near joint limits and singularities, judged by joint margin and the Jacobian's smallest singular value. & Holds: 56 and 67\%; Factory part holds. \\
\addlinespace
H2 & Insertion rate from full-task starts: the run means at the three $-x$ tilts end within 5 points of each other, at least as high as at $0^\circ$ under either threshold, and above each baseline at $-30^\circ x$: residual RL and, in planned runs, fine-tuning without the frame. & Holds so far: 4.1 points apart; about 19 points above residual RL at $-30^\circ x$. \\
\addlinespace
H3 & The twelve predictions miss by at most 5 points on average, counting misses above and below alike, and by less than predictions with $\phi=0$ (no gain) and $\phi=1$ (no drop) on the same cells; both also hold for the task-space impedance variant alone. The nine single-change cases rank by the share of the drop they avoid as by $\phi$ ($\rho>0.5$). & Planned. \\
\bottomrule
\end{tabular}
\end{table}

\begin{table}[t]
\centering
\caption{Arm configuration and success above the hole (H1)}
\label{tab:zs}
\footnotesize
\setlength{\tabcolsep}{1.9pt}
\begin{tabular}{@{}lccccccc@{}}
\toprule
 & & \multicolumn{3}{c}{mouth away from base} & \multicolumn{2}{c}{sideways} & toward \\
\cmidrule(lr){3-5}\cmidrule(lr){6-7}
Tilt & $0^\circ$ & $-15^\circ x$ & $-30^\circ x$ & $-45^\circ x$ & $-30^\circ y$ & $+30^\circ y$ & $+30^\circ x$ \\
\midrule
$\Delta q$ (rad) & 0.28 & 0.58 & 1.09 & 1.61 & 1.00 & 1.48 & 2.04 \\
$c_\parallel$ (mm/N) & 1.27 & 1.75 & 2.07 & 2.07 & 1.85 & 0.94 & 0.36 \\
Joint margin (rad) & 0.42 & 0.68 & 0.98 & 1.16 & 0.75 & 0.38 & 0.03 \\
\midrule
$\pih$ success (\%) & 91.0 & 88.6 & 73.3 & 44.5 & 53.3 & 48.1 & 12.6 \\
Factory success (\%) & 98.3 & 95.3 & 67.3 & 18.2 & 98.0 & 97.9 & 62.2 \\
\bottomrule
\end{tabular}\\[2pt]
{\raggedright\scriptsize Fixed board; $\pih$: $0^\circ$ in the tilted-hole setup, at $+30^\circ x$ including starts beyond a joint limit. Medians over 256 starts ($\Delta q$ 0.28 at $0^\circ$ from board placement); joint margin: to the nearest limit. Factory: 768 episodes, three training seeds, tilt matched in size and direction relative to the robot; at $45^\circ$ away, the 47\% of starts its reset cannot reach fail; at $45^\circ$ toward (not shown), 23.7\%, 96.4\% with relaxed limits.\par}
\end{table}

\begin{table}[t]
\centering
\caption{Correcting what the frame cannot rotate (H1, H3)}
\label{tab:decomp}
\footnotesize
\setlength{\tabcolsep}{2.5pt}
\begin{tabular}{@{}lccc@{}}
\toprule
 & $-30^\circ x$ & $-45^\circ x$ & $-30^\circ y$ \\
\midrule
\multicolumn{4}{@{}l}{\emph{At the tilted hole: success \% (share of the drop recovered, \%)}} \\
Uncorrected & 73.3 & 44.5 & 53.3 \\
Joint angles of a turned base & 78.6 (22) & 57.3 (20) & 54.2 (2) \\
Stiffness and damping & 82.7 (42) & 66.9 (36) & 66.7 (26) \\
Payload turned alone & 81.1 (34) & 67.6 (38) & 74.0 (42) \\
Joint angles, stiffness and damping, payload & 89.8 (90) & 85.6 (79) & 76.5 (48) \\
\textbf{All four corrections} & \textbf{85.1 (56)} & \textbf{81.5 (67)} & \textbf{76.6 (48)} \\
\textbf{\quad damping also matched} & \textbf{89.3 (85)} & \textbf{87.5 (85)} & \textbf{76.5 (48)} \\
Base turned$^\ddagger$ & 83.9 (55) & 67.7 (40) & 74.3 (46) \\
All rotated$^\ddagger$ & 90.7 (109) & 89.5 (99) & 89.1 (96) \\
\bottomrule
\end{tabular}\\[2pt]
{\raggedright\scriptsize From above the hole. 95\% intervals of the bold shares: 44--70, 61--74, 39--58; damping also matched: 65--110, 74--98, 39--58. Shares are relative to the $0^\circ$ run of the same setup: the tilted-hole setup (91.0\%) for the corrections, the source's task setup (89.7\%) for $\ddagger$ and injections; over 100: above that run. $\ddagger$: run at $0^\circ$, mouth 35--48\,mm lower than at the tilt, which costs at most 1.6 points. Single factors and injections: Fig.~\ref{fig:decomp}.\par}
\end{table}

\section{Preliminary Results}
\label{sec:results}

\textbf{H1: success falls with the configuration change, also when only the board moves.}
Across the 17 tilts, success falls with $\Delta q$ (Fig.~\ref{fig:tilt}a; $\rho=-0.91$). With the board placed widely, mostly beyond the training range, it falls with each episode's $\Delta q$ at all four tilts tested, including $0^\circ$ (Fig.~\ref{fig:tilt}a diamonds; Table~\ref{tab:crit}). The loss thus follows the arm's configuration, not only the tilt. With $c_\parallel$ and the payload's deflection added, $\Delta q$ keeps a negative slope at $0^\circ$, $-15^\circ x$ and $-30^\circ y$; at $-45^\circ x$, the most compliant of the four along the hole axis, these two absorb it ($\Delta q$: $-0.1$ $[-0.6,0.4]$).

\textbf{H1: with matched damping, correcting the four factors brings $-30^\circ x$ and $-45^\circ x$ within 4 points of the vertical hole.}
Together, the four corrections raise success from 73 and 45\% to 85 and 81\%, and with matched damping to 89 and 87\%, against 91\% at $0^\circ$ (Table~\ref{tab:decomp}). The arm side and the loads take most of the drop (Fig.~\ref{fig:decomp}, left): Shapley values give the joint angles with stiffness and damping 54--60\% and the loads 25--31\% at $-30^\circ x$ and $-45^\circ x$ (15 and 34\% sideways). Correcting the joint angles, stiffness and damping recovers 74 and 60\%, more than turning the base (55 and 40\%), with which the arm keeps the $0^\circ$ axial stiffness and hits the force limit about as often as at $0^\circ$; the corrected arm rarely does. The controls agree: moving the stiffness or the loads the other way lowers success below uncorrected (loads: 37, 7 and 18\%), and without the payload, the arm corrections alone recover nearly all of the drop at $-30^\circ x$ and $-45^\circ x$, as much as turning the base. At these two tilts, injecting a factor at $0^\circ$ reproduces about as much of the drop as correcting it recovers, or more (Fig.~\ref{fig:decomp}, right). At $-30^\circ x$ and $-45^\circ x$, the loss is thus mostly robot-side: at the arm, through its joint-angle input, stiffness and damping, and in the loads on the hand. At these tilts the tool is 1.6 times as compliant along the hole axis as at $0^\circ$ (Table~\ref{tab:zs}), and gravity on the payload, axial at $0^\circ$, acts partly across the hole.

At $-30^\circ y$, the four corrections raise success from 53 to 77\%. Sideways, the symmetry also breaks after the release, when the part is no longer held and the source relies on gravity to seat it: injected at $0^\circ$, the tilt's loads reproduce 23\% of the drop while held but 50\% with the part's weight turned also after release.

The Factory policy, to which only the joint limits apply, loses only near limits and singularities. It keeps 98\% sideways, where $\pih$ halves (Table~\ref{tab:zs}). Relaxed limits restore it only at $45^\circ$ toward the robot, where the arm starts at a limit. Elsewhere, it loses where the Jacobian's smallest singular value falls from 0.19 to 0.06--0.11 (0.18 sideways).

\textbf{H2: fine-tuned in the frame, the $-x$ tilts end close together and at least match the vertical hole.}
The $-x$ tilts rise from 18--47\% to 75--79\% in insertion rate and end 4.1 points apart (Fig.~\ref{fig:tilt}b). All three end above the vertical hole under either threshold, and at $-30^\circ x$ about 19 points above residual RL in both settings. Full-task success rises from 12--44\% to 73--76\% (35.4 to 75.5\% at $-30^\circ x$). The vertical hole, where the arm is stiffer along the hole axis ($c_\parallel$, Table~\ref{tab:zs}), lags with the 11\,N threshold (62.8\%); with the threshold scaled to 18\,N, it reaches 72.7\%, 2.6--6.7 points below the $-x$ tilts.

\section{Expected Insights and Planned Experiments}
\label{sec:plan}

For our policy, Sec.~\ref{sec:results} traces most of the loss at $30^\circ$ and $45^\circ$ away from the base to three robot-side choices: reading joint angles, joint impedance, and training against a load whose direction the frame cannot turn. We will test whether the corrections predict what changing these choices in training saves (H3), and check the main comparisons on a real arm.

(1)~\emph{Designs (H3).} If the predictions for the four retrained variants hold, the corrections tell, before retraining, how much a design choice such as dropping joint-angle inputs or using isotropic task-space impedance would save at a tilt. If a variant loses more than predicted, its own decomposition shows whether other factors cost it more, the design adds a robot-side loss, or the correction overstates the factor; if it loses less, the correction understates the factor.

(2)~\emph{Baselines (H2).} We add baselines without the frame at the same budget: fine-tuning with added roll and pitch actions, from the source and after cloning the 3.5\% of source trajectories that succeed when moved to the tilt~\cite{mandlekar2023mimicgen,zhou2025reinforcegen}. If fine-tuning in the frame over random tilts, with the single-tilt runs' combined budget, comes within 5 points of each tilt's own run in insertion, one policy covers the tilts. We also add 18\,N runs at $-30^\circ x$ and more at $0^\circ$: if the higher threshold's logit gain at $-30^\circ x$ lies within the $0^\circ$ gain's interval, the penalty limits fine-tuning everywhere; if it is smaller, the threshold should scale with $1/c_\parallel$.

(3)~\emph{Real arm (H1, H2).} We first calibrate the simulated payload: we measure how far known loads move the tool at the $0^\circ$, $-30^\circ x$, $-45^\circ x$ and $-30^\circ y$ starts; the predictions below keep the payload if the tool moves at least half as far as simulated, and use 0\,N otherwise. The main test is the full task on a $-30^\circ x$ wedge, $\pih$ against the policy fine-tuned there (35.4 vs.\ 75.5\% full-task success in simulation). Fine-tuning transfers if it raises success (one-sided test, 5\%); if it does not, the above-hole test helps locate why. Above the hole, we run $\pih$ on a flat board and on $-30^\circ x$ and $-45^\circ x$ wedges, with the hole located the same way on all three. The drop that H1 explains appears if each wedge succeeds less often than the flat board (one-sided, 5\%), and appears as simulated unless a real 95\% interval, at most 15 points in half-width, excludes the simulated rate. If the drops appear as simulated, simulation predicts what a retrofit loses; otherwise, a miss on the flat board points to the simulation as a whole, and a miss on the wedges alone to configuration effects the simulation lacks.

\textbf{Conclusion.} The hole frame turns the task but not the robot. For our policy, which reads joint angles under joint impedance with a load on its hand, the symmetry breaks at the arm and in the loads; correcting both, with matched damping, brings $30^\circ$ and $45^\circ$ away from the base within 4 points of the vertical hole. Fine-tuning in the frame lifts the insertion rate at these tilts to at least that of the source fine-tuned the same way at $0^\circ$.

\clearpage
\setlength{\bibsep}{0.5pt plus 0.3pt}
\renewcommand{\bibfont}{\footnotesize}
\bibliographystyle{plainnat}
\bibliography{references_v9}

\end{document}